\documentclass[letterpaper, 10 pt, conference]{ieeeconf}  

\IEEEoverridecommandlockouts                              
\usepackage{duckuments}

\usepackage{algorithm}
\usepackage{balance}

\usepackage{graphicx}
\usepackage{epstopdf}

\usepackage{colortbl}

\usepackage{subcaption}

\usepackage{todonotes}

\usepackage{amsmath,amssymb,amsfonts}
\usepackage{xcolor}
\usepackage{bm}
\usepackage{multirow}

\usepackage{cite}

\usepackage{amsmath}

\usepackage{hyperref} 
\hypersetup{
    colorlinks=true,
    linkcolor=black,
    urlcolor=blue,
}

\usepackage{amsthm}
\newtheorem{remark}{Remark}

\usepackage{scalerel}
\def\shrinkage{-2.4mu}
\def\vecsign#1{\rule[1.388\LMex]{\dimexpr#1-2.5pt}{.36\LMpt}%
  \kern-6.0\LMpt\mathchar"017E}
\def\dvecsign#1{\rule{0pt}{7\LMpt}\smash{\stackon[-1.989\LMpt]{%
  \SavedStyle\mkern-\shrinkage\vecsign{#1}}%
  {\rotatebox{180}{$\SavedStyle\mkern-\shrinkage\vecsign{#1}$}}}}
\def\dvec#1{\ThisStyle{\setbox0=\hbox{$\SavedStyle#1$}%
  \def\useanchorwidth{T}\stackon[-4.2\LMpt]{\SavedStyle#1}{\,\dvecsign{\wd0}}}}
\usepackage{stackengine,amsmath}
\stackMath

\title{\LARGE \bf
Hierarchical Reduced-Order Model Predictive Control for \\ Robust Locomotion on Humanoid Robots}

\author{Adrian~B.~Ghansah$^{1}$, Sergio~A.~Esteban$^{2}$, Aaron~D.~Ames$^{1,2}$
\thanks{$^{1}$A. B. Ghansah, and A. D. Ames are with the Department of Control and Dynamical Systems, California Institute of Technology, Pasadena, CA 91125, USA, {\tt\small \{aghansah, ames\}@caltech.edu}\newline
$^{2}$S. A. Esteban and A. D. Ames are with the Department of Mechanical and Civil Engineering, California Institute of Technology, Pasadena, CA 91125, USA, {\tt\small \{sesteban, ames\}@caltech.edu}
}%
\thanks{This research is supported by Technology Innovation Institute (TII).}
}

\begin{document}

\maketitle
\thispagestyle{empty}
\pagestyle{empty}

\begin{abstract}
    As humanoid robots enter real-world environments, ensuring robust locomotion across diverse environments is crucial.  This paper presents a computationally efficient hierarchical control framework for humanoid robot locomotion based on reduced-order models---enabling versatile step planning and incorporating arm and torso dynamics to better stabilize the walking. At the high level, we use the step-to-step dynamics of the ALIP model to simultaneously optimize over step periods, step lengths, and ankle torques via nonlinear MPC. The ALIP trajectories are used as references to a linear MPC framework that extends the standard SRB-MPC to also include simplified arm and torso dynamics. We validate the performance of our approach through simulation and hardware experiments on the Unitree G1 humanoid robot. In the proposed framework the high-level step planner runs at 40 Hz and the mid-level MPC at 500 Hz using the onboard mini-PC. Adaptive step timing increased the push recovery success rate by 36\%, and the upper body control improved the yaw disturbance rejection. We also demonstrate robust locomotion across diverse indoor and outdoor terrains, including grass, stone pavement, and uneven  gym mats.
\end{abstract}

\section{Introduction}\label{sec:introduction}
%

Deployment of humanoid robots in real-world scenarios demands controllers that are both robust and reliable. The high-degree-of-freedom (DOF), hybrid, and underactuated dynamics, combined with the need to interact with diverse environments make reliable control of humanoids a significant challenge. Planning over the full robot state is computationally expensive, often rendering algorithms unsuitable for hardware deployment where the computational power may be limited and real-time control is essential. The hybrid dynamics necessitate careful reasoning about contact locations and timings, which nominally leads to large and challenging mixed-integer optimization problems.

The hybrid zero dynamics (HZD) framework \cite{westervelt2003hybrid} is based on finding stable periodic orbits for the full-order hybrid dynamical system through offline trajectory optimization, and then to stabilize the trajectories online. The method has been demonstrated across various different platforms \cite{reher2016realizing, reher2019dynamic, ghansah2023humanoid}. Despite its strong theoretical foundation, the need for offline trajectory optimization makes the method less suitable in dynamic scenarios where online replanning is required. 

\begin{figure}
    \centering
    \href{https://vimeo.com/1110476518}
    {
    \includegraphics[width=0.98\linewidth]{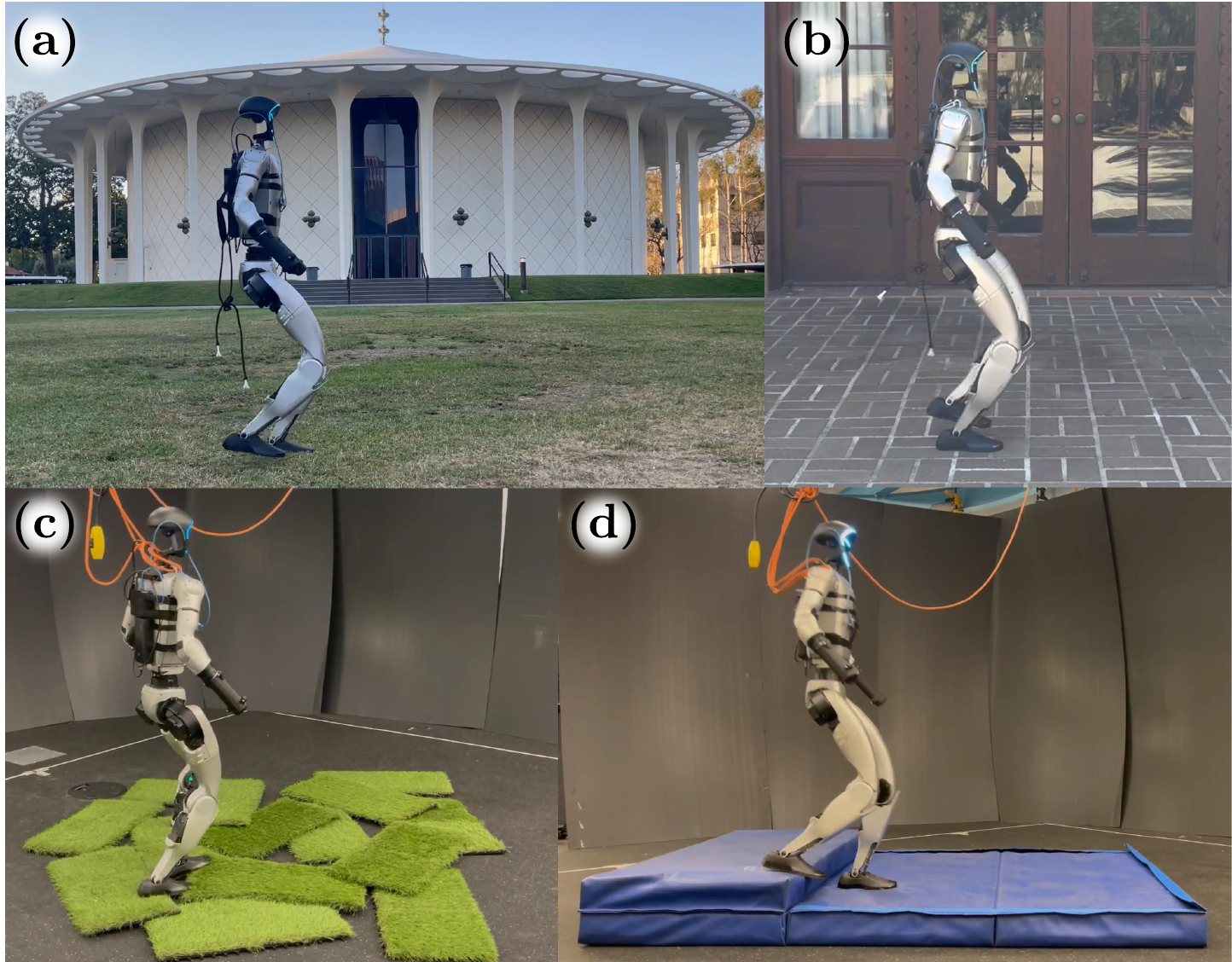}
    }
    \caption{The G1 humanoid robot traversing a variety of terrains using our reduced-order model predictive control framework.}
    \label{fig:hero}
    \vspace{-5mm}
\end{figure}

Approaches involving whole-body model predictive control (MPC) benefit from future predictions to stabilize the underactuated humanoid dynamics.
The use of whole-body MPC can lead to rich and complex behaviors \cite{dantec2022whole, esteban2025reduced}. While whole-body MPC has been successfully deployed on hardware \cite{khazoom2024tailoring}, solving the associated optimization problems are often computationally expensive, extensive tuning may be required, and it may be subject to local minima. 

An alternative to using the full-order model is to instead employ a reduced-order model (ROM), such as the linear inverted pendulum (LIP) \cite{kajita20013d}, hybrid LIP (HLIP) \cite{xiong20223}, angular LIP (ALIP) \cite{gong2020angular}, or spring-loaded inverted pendulum (SLIP) \cite{blickhan1989spring}. A ROM is a reduced representation of the full-order model that captures the underlying core dynamics of locomotion. Stabilizing the ROM and embedding it within the full-order system stabilizes the whole-body dynamics. The lower state dimension and simplified dynamics makes ROM control approaches computationally efficient and well suited for real-time control. However, controllers using ROMs are less expressive as they nominally do not consider the full system dynamics such as arm swinging or turning.
 
\begin{figure*}[t]
    \centering
    \includegraphics[width=1.0 \linewidth]{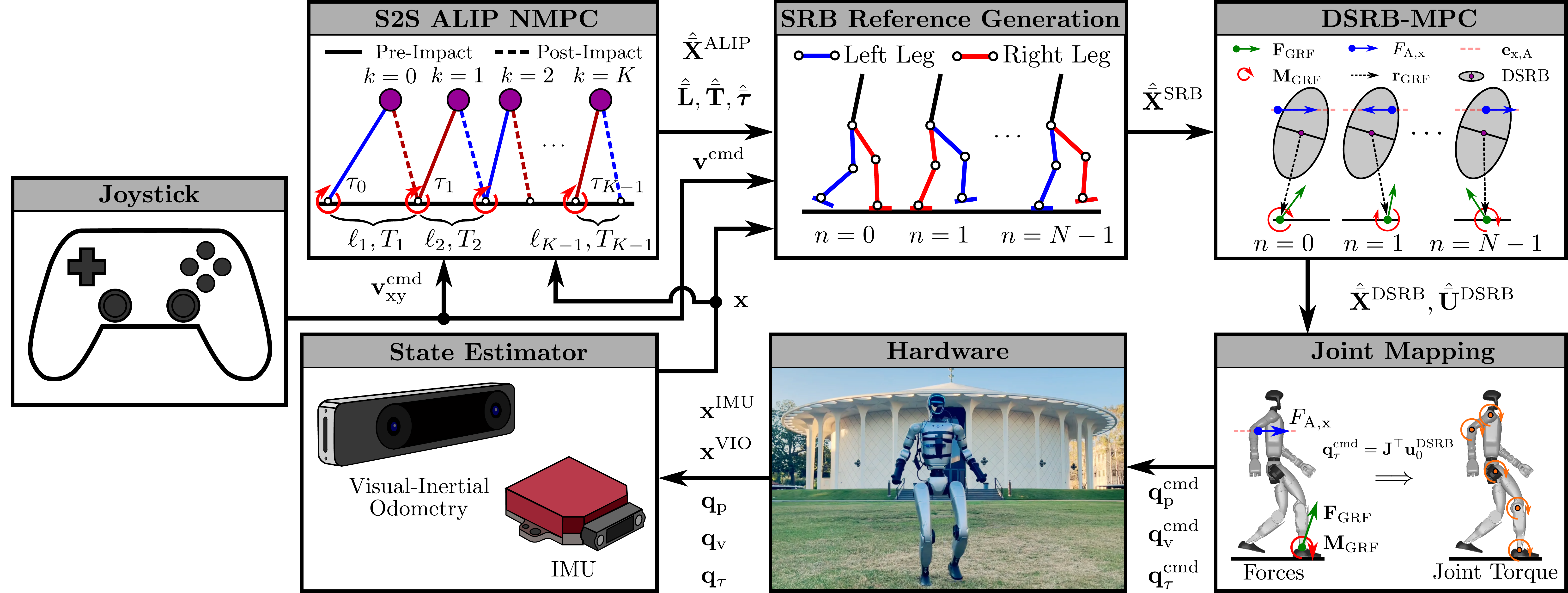}
    \caption{In the high-level planner, NMPC is used on the S2S dynamics of the ALIP model to generate a step plan. The step plan is then converted into a SRB reference trajectory. The reference trajectory is passed to a linear MPC problem that optimizes ground reaction forces and the upper body outputs. The output references together with the forces and torques from the MPC is passed to joint mapper which converts all the references into joint commands that are sent to the robot.}
    \label{fig:control_architecture}
    \vspace{-5mm}
\end{figure*}

\subsection{Related Work}

\subsubsection{Foot Placement Strategies}

Several works have studied use of ROMs to stabilize bipedal robots. In \cite{xiong2021global}, MPC is used to optimize over step lengths based on the HLIP model to achieve global position tracking. The authors in \cite{gibson2022terrain} use MPC on the ALIP model to find the optimal step length to track a commanded velocity. In \cite{dosunmu2023stair}, a reparameterization of the ALIP model is used and include ankle torques to achieve stair-climbing. In these works, step periods are fixed to a constant value, leading to simpler optimization problems.

In addition to planning foot placement locations, incorporating step durations for optimization has also been investigated. In \cite{kim2023model}, a hierarchical control strategy is presented where a high-level MPC based on the LIP Flywheel (LIPF) model optimizes over ankle control, hip control, and step lengths, followed by a subsequent layer that optimizes step durations. The authors in \cite{choe2023seamless} propose a nonlinear MPC (NMPC) controller based on the LIPF model and the divergent component of motion (DCM) error that optimizes over ankle control, hip control, step lengths, and step periods simultaneously. A heuristic stepping strategy is proposed in \cite{li2024adapting} where the nominal stepping period is increased based on the evolution of unstable states and body attitude changes in a high-level NMPC problem.

\subsubsection{Upper Body Control}
A key advantage of whole-body control is that it inherently reasons about torso and arm motion which has shown to give rise to emergent behaviors \cite{dai2014whole, khazoom2024tailoring, esteban2025reduced}. In contrast, several popular inverted pendulum variants exclude upper-body dynamics by design. The LIPF model can be used to strategize about upper body motion to stabilize the robot, as in \cite{kim2023model}. Model Hierarchy Predictive Control (MHPC) has also shown promise in enhancing robustness through arm control \cite{khazoom2022humanoid}. The authors in \cite{xing2010arm} generate arm movement for a robot using zero moment point walking to find arm swinging motions to avoid torsional rotation of the feet around the z-axis. 

\subsection{Contributions}

In this paper, we propose a hierarchical control framework for humanoid locomotion that aims at bridging the gap between fast, simplified ROM-based control and more complex whole-body control. At the high-level, we use NMPC on the step-to-step (S2S) dynamics of the ALIP model to efficiently optimize over step lengths, step periods, and ankle torques at 40 Hz. The resulting plans are then used to generate reference trajectories for a linear Single Rigid Body (SRB) MPC. We then show how the SRB model can be decomposed to incorporate arm and torso dynamics, while still keeping the dynamics constraints in the MPC linear. We demonstrate that adaptive stepping improves robustness and that upper-body control helps mitigate unwanted yaw rotations. An overview of the framework is shown in Fig. \ref{fig:control_architecture}.

The hierarchical control structure allows the high-level planner to solve the nonconvex step length and step period optimization at a lower rate, while the mid-level controller is formulated as a linearly constrained MPC that can be executed at high rates. This separation ensures predictable solve times at the mid level while enabling more versatile step planning at the high level.

Our main contributions are:
\begin{enumerate}
\item A high-level NMPC formulation based on ALIP S2S dynamics that jointly optimizes over step lengths, step durations, and ankle torques.
\item A decomposition of the SRB model to include arm and torso dynamics, while maintaining linear dynamics.
\item Extensive simulation and hardware experiments to demonstrate the robustness of the proposed approach.
\end{enumerate}

The remainder of the paper is organized as follows. Section \ref{sec:background} introduces the classical SRB model and the ALIP S2S dynamics. Section \ref{sec:method} details the NMPC formulation, the mapping from S2S plans to SRB references, and the decomposition of the SRB model to also include torso and arm movement. Section \ref{sec:experiments} presents simulation and hardware results, and in Section \ref{sec:conclusion} we offer some concluding remarks. A video of the experiments is available at:\newline \url{https://vimeo.com/1110476518}

\section{Preliminaries}\label{sec:background}


\subsection{Whole Body Dynamics}
For a 3D humanoid robot we denote the configuration space as
$
    \mathbf q
    \in  \mathcal{Q}
$
and the velocities as
$
    \mathbf v
    \in  \mathcal{V}
$
.
The whole body state of the robot is then represented as $\mathbf x~=~[\mathbf q^\top, \mathbf v^\top]^{\top} \in \mathcal{X} \triangleq \mathcal{Q} \times \mathcal{V}$.
%
%
\subsection{Single Rigid Body Dynamics}
The SRB dynamics of the robot can be obtained using Newton-Euler dynamics
\begin{equation}
\label{eq:newton_euler}
\scalebox{0.9}{$
    \begin{bmatrix}
        \mathbf{F} \\
        \mathbf{M}
    \end{bmatrix}
    =
    \begin{bmatrix}
        m_\textnormal{COM} \boldsymbol{I}_{3 \times 3} & \mathbf{0}_{3 \times 3} \\
        \mathbf{0}_{3 \times 3} & \mathbf{I}_\textnormal{COM}
    \end{bmatrix}
    \begin{bmatrix}
        \dot{\mathbf{v}}^\textnormal{COM} \\
        \dot{\boldsymbol{\omega}}^\textnormal{COM}
    \end{bmatrix}
    +
    \begin{bmatrix}
        \mathbf{0}_{3 \times 1} \\
        \left( \boldsymbol{\omega}^\textnormal{COM} \right)^{\times} \mathbf{I}_\textnormal{COM} \boldsymbol{\omega}^\textnormal{COM}
    \end{bmatrix}.
$}
\end{equation}
$\mathbf{F} \in \mathbb{R}^3$ and $\mathbf{M} \in \mathbb{R}^3$ are the external forces and moments acting on the robot, represented in the inertial frame, $m_\textnormal{COM} \in \mathbb{R}$ and $\mathbf{I}_\textnormal{COM} \in \mathbb{R}^{3 \times 3}$ are the robot's mass and inertia around the COM in the inertial frame. $\mathbf{v}^\textnormal{COM}$,  $\boldsymbol{\omega}^\textnormal{COM} \in \mathbb{R}^3$ are the linear and angular velocity of the COM frame relative to the inertial frame.
We represent the full SRB state as
\begin{equation}
    \label{eq:srb_state}
    \mathbf{x}^\textnormal{SRB} 
    = 
    \begin{bmatrix}
    \textbf{p}^\textnormal{COM} & \boldsymbol{\Theta}^\textnormal{COM} & \textbf{v}^\textnormal{COM} & \boldsymbol{\omega}^\textnormal{COM} & g
    \end{bmatrix},
\end{equation}
where $\mathbf{p}^\textnormal{COM} \in \mathbb{R}^3$ and $\boldsymbol{\Theta}^\textnormal{COM} = [\phi \ \theta \ \psi]^\top \in \mathfrak{so}(3)$ is the COM frame position and orientation relative to the inertial frame. $\phi$, $\theta$, and $\psi$ denote the roll, pitch, and yaw. We include the gravity vector in the state as it allows us to write the dynamics in affine form without adding a constant term 
\begin{equation}
    \label{eq:continous_time_dynamics}
    \dot{\mathbf{x}}^\textnormal{SRB} = \mathbf{A}^\textnormal{SRB}(\mathbf{x}^\textnormal{SRB}) \mathbf{x}^\textnormal{SRB} + \mathbf{B}^\textnormal{SRB}(\mathbf{x}^\textnormal{SRB}, \mathbf{p}^\textnormal{STF}) \mathbf{u}^\textnormal{SRB}.
\end{equation}
In \eqref{eq:continous_time_dynamics}, $\mathbf{u}^\textnormal{SRB} = [\mathbf{F}_\textnormal{GRF,L}^\top \ \mathbf{M}_\textnormal{GRF,L}^\top \ \mathbf{F}_\textnormal{GRF,R}^\top \ \mathbf{M}_\textnormal{GRF,R}^\top]^\top$ are the external forces and torques acting on the robot at the feet, and $\mathbf{p}^\textnormal{STF}$ is the stance foot (STF) location in the inertial frame. The expressions for $\mathbf{A}^\textnormal{SRB}(\mathbf{x}^\textnormal{SRB})$ and $\mathbf{B}^\textnormal{SRB}(\mathbf{x}^\textnormal{SRB}, \mathbf{p}^\textnormal{STF})$ are omitted here for brevity but can be found in \cite{li2021force}. 


\subsection{Step-to-Step ALIP Dynamics}

The 3D ALIP model is composed of two LIP models, one in the sagittal plane and one in the frontal plane, where the states are given by $\mathbf{x}^\textnormal{ALIP} = [p_\textnormal{x} \ L_\textnormal{y} \ p_\textnormal{y} \ L_\textnormal{x}]^{\top}$. $p_\textnormal{x/y}$ is the horizontal COM x/y position relative to the stance foot, and $L_\textnormal{y/x}$ is the angular momentum around the stance foot about the y-axis and the x-axis \cite{gong2020angular}. The nominal ALIP model assumes a fixed height, $p_\textnormal{z}$, and that the angular momentum around the pre-impact swing foot is conserved during foot switching. Furthermore, if we have ankle torque $\tau$ available, the continuous dynamics for the actuated ALIP model during a step are given by:
\begin{equation}
\label{eq:alip_linear_dynamics}
\scalebox{0.8}{$
    \begin{bmatrix}
        \dot{p}_\textnormal{x} \\
        \dot{L}_\textnormal{y} \\
        \dot{p}_\textnormal{y} \\
        \dot{L}_\textnormal{x}
    \end{bmatrix}
    =
    \underbrace{
    \begin{bmatrix}
        0 & \frac{1}{m_\textnormal{COM} p_\textnormal{z}} & 0 & 0 \\
        m_\textnormal{COM} g & 0 & 0 & 0 \\
        0 & 0 & 0 & -\frac{1}{m_\textnormal{COM} p_\textnormal{z}} \\
        0 & 0 & -m_\textnormal{COM} g & 0 \\    
    \end{bmatrix}
    }_{\mathbf{A}^\textnormal{ALIP}}
    \underbrace{
    \begin{bmatrix}
        p_\textnormal{x} \\
        L_\textnormal{y} \\
        p_\textnormal{y} \\
        L_\textnormal{x}
    \end{bmatrix}
    }_{\mathbf{x}^\textnormal{ALIP}}
    +
    \underbrace{
    \begin{bmatrix}
        0 & 0 \\
        1 & 0\\
        0 & 0 \\
        0 & 1 
    \end{bmatrix}
    }_{{\mathbf{B}^\textnormal{ALIP}}}
    \underbrace{
    \begin{bmatrix}
        \tau_\textnormal{y} \\
        \tau_\textnormal{x}
    \end{bmatrix}
    }_{{\boldsymbol{\tau}}}
$}
\end{equation}

In \eqref{eq:alip_linear_dynamics}, $\tau_\textnormal{y}$ and $\tau_\textnormal{x}$ are the ankle torques in the sagittal plane and frontal plane. The continuous dynamics during a single support phase can be compactly written as
\begin{equation}
    \label{eq:lti_dynamics}
    \dot{\mathbf{x}}^\textnormal{ALIP}(t) = \mathbf{A}^\textnormal{ALIP} \mathbf{x}^\textnormal{ALIP}(t) + \mathbf{B}^\textnormal{ALIP} \boldsymbol{\tau}(t).
\end{equation}
Given an initial state at time $t = t_0$, a constant $\boldsymbol{\tau}$, and $T > 0$, we can flow the dynamics forward in time, and we have that the state at time $t = t_0 + T$ is given by
\begin{equation}
    \label{eq:nonlinear_forward_propagation}
    \mathbf{x}^\textnormal{ALIP}(t_0 + T) = \boldsymbol{\Phi}(T)  \mathbf{x}^\textnormal{ALIP} (t_0) + \boldsymbol{\Gamma}(T) \boldsymbol{\tau},
\end{equation}
where $\boldsymbol \Phi$ and $\boldsymbol 
 \Gamma$ are given by
\begin{align}
    \boldsymbol{\Phi}(T) &= \mathbf{e}^{\mathbf{A}^\textnormal{ALIP} T}, \\
    \boldsymbol{\Gamma}(T) &= \left( \int_{t_0}^{t_0 + T} \mathbf{e}^{\mathbf{A}^\textnormal{ALIP}(t_0 + T - s)} \, ds \right) \mathbf{B}^\textnormal{ALIP}.
\end{align}


Let the state immediately before step $k$, i.e. the pre-impact state, be denoted by $\left(\mathbf{x}^\textnormal{ALIP}_k\right)^-$ and the state right after impact at step $k$, i.e. the post impact state, be denoted by $\left(\mathbf{x}^\textnormal{ALIP}_{k+1}\right)^+$. Given a step length $\boldsymbol{\ell} = [\ell_\textnormal{x} \ \ell_\textnormal{y}]^\top$, we have that the pre-impact and post-impact states are related as follows
\begin{equation}
    \label{eq:alip_reset_map}
    \left(\mathbf{x}^\textnormal{ALIP}_{k+1}\right)^+
    = 
    \left(\mathbf{x}^\textnormal{ALIP}_k\right)^-
    + 
    \mathbf{B}_d
    \boldsymbol{\ell}_k,
\end{equation}
where $\mathbf{B}_d = [-1 \ 0 \ -1 \ 0]^{\top}$. 
Furthermore, from \eqref{eq:nonlinear_forward_propagation} we have that the pre-impact state at step $k$ can be obtained from the post-impact state at step $k$. We can then write: 
\begin{equation}
    \label{eq:post_impact_to_pre_impact}
    \left(\mathbf{x}^\textnormal{ALIP}_k \right)^-  = \boldsymbol{\Phi}(T_k)  \left(\mathbf{x}^\textnormal{ALIP}_k \right)^+ + \boldsymbol{\Gamma}(T_k) \boldsymbol{\tau}_k.
\end{equation}
Inserting \eqref{eq:post_impact_to_pre_impact} into \eqref{eq:alip_reset_map} gives us the S2S pre-impact dynamics
\begin{equation}
    \label{eq:nonlinear_s2s_dynamics}
    \mathbf{x}^\textnormal{ALIP}_{k+1} = \boldsymbol{\Phi}(T_k)  \mathbf{x}^\textnormal{ALIP}_{k} + \boldsymbol{\Gamma}(T_k) \boldsymbol{\tau}_k + \mathbf{B}_\textnormal{d} \boldsymbol{\ell}_k,
\end{equation}
where $\mathbf{x}^\textnormal{ALIP}_k$ is the pre-impact state at step $k$.


\subsection{Single Rigid Body Model Predictive Control}\label{sec:background:linear_mpc}

Given the current full-order state, $\mathbf{x}_0$, a SRB reference trajectory, $\hat{\bar{\mathbf{X}}}^\textnormal{SRB} = \left\{ \bar{\mathbf{x}}_{n}^\textnormal{SRB} \right\}_{n = 1}^{N}$, a stance foot trajectory, $\hat{\bar{\mathbf{P}}}^\textnormal{STF} = \left\{ \bar{\mathbf{p}}_{n}^{\textnormal{STF}} \right\}_{n = 0}^{N-1}$, and a MPC time discretization sequence $ \Delta \hat{\mathbf{T}} = \left\{ \Delta t_n \right\}_{n = 0}^{N-1}$, we can linearize the SRB dynamics around the references at each timestep $n$ in the horizon to obtain the linearized discrete dynamics
\begin{equation}
    \mathbf{x}_{n+1}^\textnormal{SRB} = \mathbf{A}^\textnormal{SRB}_n \mathbf{x}_{n}^\textnormal{SRB} + \mathbf{B}^\textnormal{SRB}_{n} \mathbf{u}_{n}^\textnormal{SRB},
\end{equation}
where $\mathbf{A}^\textnormal{SRB}_{n}$ and $\mathbf{B}^\textnormal{SRB}_{n}$ are given by
\begin{equation}
    \mathbf{A}^\textnormal{SRB}_{n} = \boldsymbol{I}_{13 \times 13} + {\Delta t}_{n} \mathbf{A}^\textnormal{SRB }(\bar{\mathbf{x}}^\textnormal{SRB}_n) 
\end{equation}
\begin{equation}
    \mathbf{B}^\textnormal{SRB}_{n} = {\Delta t}_{n} \mathbf{B}^\textnormal{SRB}(\bar{\mathbf{x}}^\textnormal{SRB}_n, \bar{\mathbf{p}}^\textnormal{STF}_n).
\end{equation}
At $n = 0$ we discretize around the current state, $\mathbf{x}_0$. Using the linearized discrete SRB dynamics and reference trajectory we can then formulate the following linear MPC problem
\begin{equation}
\label{eq:srb_mpc}
\begin{aligned}
    \min_{\mathbf{x}^\textnormal{SRB}, \mathbf{u}^\textnormal{SRB}} & \sum_{n = 0}^{N-1} \|{\mathbf{x}}_{n+1}^\textnormal{SRB} - \bar{\mathbf{x}}_{n+1}^\textnormal{SRB} \|_{\mathbf{Q}^\textnormal{SRB}} + \| {\mathbf{u}_{n}^\textnormal{SRB}} \|_{\mathbf{R}^\textnormal{SRB}} \\
    \text{s.t.} \quad & \mathbf{x}_{n+1}^\textnormal{SRB} = \mathbf{A}^\textnormal{SRB}_{n} \mathbf{x}_{n}^\textnormal{SRB} + \mathbf{B}^\textnormal{SRB}_{n} \mathbf{u}_{n}^\textnormal{SRB}, n \in \{0, ..., N-1 \} \\
    & \mathbf{C}_n^\textnormal{SRB} \mathbf{u}_n^\textnormal{SRB} \leq \mathbf{c}_n^\textnormal{SRB}, \hspace{5.7em} n \in \{0, ..., N-1 \} \\
    & \mathbf{D}_n^\textnormal{SRB} \mathbf{u}_{n}^\textnormal{SRB} = \mathbf{0}, \hspace{6.95em} n \in \{0, ..., N-1 \}
\end{aligned}
\end{equation}

where $\mathbf{C}_n^\textnormal{SRB} \mathbf{u}_n^\textnormal{SRB} \leq \mathbf{c}_n^\textnormal{SRB}$ and $\mathbf{D}_n^\textnormal{SRB} \mathbf{u}_{n}^\textnormal{SRB} = \mathbf{0}$ denote the inequality and equality constraints at node $n$. The inequality constraints can include constraints such as friction cone constraints and force limits, while the equality constraint are often used to ensure zero force and moments from the swing foot. The MPC problem in \eqref{eq:srb_mpc} is a quadratic program (QP), which is suitable for real-time control.


\section{Method}\label{sec:method}


This section outlines our hierarchical control framework. Section \ref{sec:method:nmpc} introduces the high-level NMPC planner, which computes optimal step periods and lengths based the ALIP S2S dynamics. Section \ref{sec:method:reference_generation} describes how the S2S trajectories are converted into SRB references. In Section \ref{sec:method:arm_mpc}, we decompose the SRB-MPC formulation to include arm movement and torso rotation. In Section \ref{sec:method:low_level_control}, we show how the optimization solutions are mapped to joint commands.

\subsection{Step-to-Step ALIP Nonlinear Optimization Problem}\label{sec:method:nmpc}

\subsubsection{Nonlinear MPC Formulation}\label{sec:method:nmpc:formulation}

The objective of the high-level planner is to find optimal step length and step period trajectories based on the current state of the robot to make it walk at a commanded velocity with a desired step period. Optimizing step lengths and step periods over the full-order dynamics is computationally expensive, hence we use the S2S ALIP dynamics from \eqref{eq:nonlinear_s2s_dynamics} instead. Based on the ALIP dynamics we formulate the following optimization problem where we attempt to find the optimal step length and step periods for a horizon of $K$ steps:
\begin{equation}
\label{eq:nonlinear_mpc}
\begin{aligned}
    \min_{\mathbf{x}^\textnormal{ALIP}, \boldsymbol{\ell}, \mathbf{T}, \boldsymbol{\tau}} \quad
    & \sum_{k = 0}^{K-1} 
    \| \tilde{\mathbf{x}}^\textnormal{ALIP}_{k+1} \|_{\mathbf{Q}_\textnormal{x}} 
    + \| \tilde{{\boldsymbol{\ell}}}_{k} \|_{\mathbf{R}_{\ell}}
    + \| \tilde{{T}}_{k} \|_{{R}_T} 
    + \| {\boldsymbol{\tau}}_{k} \|_{\mathbf{R}_{\tau}} \\[0.5em]
    \text{s.t.} \quad 
    & \textnormal{Eq.}~\eqref{eq:nonlinear_s2s_dynamics}, \hspace{8.0em} k \in \{1, \dots, K{-}1\} \\
    & {\mathbf{x}}^\textnormal{ALIP}_\textnormal{lb} \leq \mathbf{x}^\textnormal{ALIP}_k \leq {\mathbf{x}}^\textnormal{ALIP}_\textnormal{ub},
      \hspace{2.1em} k \in \{1, \dots, K\} \\
    & {\boldsymbol{\ell}}_\textnormal{lb} \leq \boldsymbol{\ell}_k \leq {\boldsymbol{\ell}}_\textnormal{ub},
      \hspace{5.5em} k \in \{0, \dots, K{-}1\} \\
    & {T}_\textnormal{lb} \leq T_k \leq {T}_\textnormal{ub},
      \hspace{5.2em} k \in \{0, \dots, K{-}1\} \\
    & {\boldsymbol{\tau}}_\textnormal{lb} \leq \boldsymbol{\tau}_k \leq {\boldsymbol{\tau}}_\textnormal{ub},
      \hspace{5.4em} k \in \{0, \dots, K{-}1\} \\
    & \mathbf{x}^\textnormal{ALIP}_1 = 
    \boldsymbol{\Phi}(T_0 {-} t_\textnormal{curr}) \, \mathbf{x}^\textnormal{ALIP}_\textnormal{0} 
    + \boldsymbol{\Gamma}(T_0 {-} t_\textnormal{curr}) \, \boldsymbol{\tau}_0
\end{aligned}
\end{equation}

In \eqref{eq:nonlinear_mpc}, the tilde $\tilde{(\cdot)}$ denotes the difference between the actual value and the desired value. This gives us $\tilde{\mathbf{x}}^\textnormal{ALIP}_k = \mathbf{x}^\textnormal{ALIP}_k - {\mathbf{x}}^\textnormal{ALIP,des}_k$, $\tilde{\boldsymbol{\ell}}_k = \boldsymbol{\ell}_k - {\boldsymbol{\ell}}_k^\textnormal{des}$, and $\tilde{T}_k = T_k - {T}_k^\textnormal{des}$. $\mathbf{x}^\textnormal{ALIP}_\textnormal{0}$ is the current ALIP state of the robot, and $t_\textnormal{curr}$ is the time since the beginning of the current step. For the NLP in \eqref{eq:nonlinear_mpc} we have that the cost is quadratic and all the constraints are linear, except for the dynamics constraints from \eqref{eq:nonlinear_s2s_dynamics} where  nonlinearity is introduced by considering optimization of $T_k$. 

\subsubsection{Obtaining Desired S2S Parameters}
Given a velocity command $\mathbf{v}^\textnormal{cmd} = (v_\textnormal{x}^\textnormal{cmd}, v_\textnormal{y}^\textnormal{cmd}, \omega_\textnormal{z}^\textnormal{cmd})$, and a desired step period $
{T}^\textnormal{des}$, we calculate the desired step lengths as
\begin{align}
    \label{eq:l_x}
    {\ell}_\textnormal{x}^{\textnormal{des}} = v_\textnormal{x}^\textnormal{cmd} {T}^\textnormal{des},\\
    \label{eq:l_y}
    \ell_\textnormal{y}^{\textnormal{des}} = v_\textnormal{y}^\textnormal{cmd} {T}^\textnormal{des} + \ell_\textnormal{y}^\textnormal{offset} \gamma,
\end{align}
where $\ell_\textnormal{y}^\textnormal{offset}$ is the nominal desired step width, and $\gamma \in \{-1, 1\}$ indicates the step direction based on the stance foot. 

We decide the desired pre-impact position and velocity based on the HLIP model as discussed in \cite{xiong20223}, where we set the single support step period to $T_\textnormal{SSP} = {T}^\textnormal{des}$, and we set the double support step period to zero, i.e. $T_\textnormal{DSP} = 0$. For the x-direction, we use a period-1 orbit, and for the y-direction we use a period-2 orbit. This gives us:
\begin{equation}
    \begin{bmatrix}
    {\mathbf{x}}^\textnormal{HLIP,des} \\
    {\mathbf{y}}^\textnormal{HLIP,des}
    \end{bmatrix}
    =
    \begin{bmatrix}
    \textnormal{Period1}(T_\textnormal{SSP}, T_\textnormal{DSP}, {p}_\textnormal{z}^\textnormal{des}, v_\textnormal{x}^\textnormal{cmd}) \\
    \textnormal{Period2}(T_\textnormal{SSP}, T_\textnormal{DSP}, {p}_\textnormal{z}^\textnormal{des}, v_\textnormal{y}^\textnormal{cmd}, \ell_\textnormal{y}^\textnormal{offset})
    \end{bmatrix},
\end{equation}
where $\bar{\mathbf{x}}^\textnormal{HLIP,des} = [{p}_\textnormal{x}^\textnormal{HLIP,des} \ {v}_\textnormal{x}^\textnormal{HLIP,des}]^\top$ and ${\mathbf{y}}^\textnormal{HLIP,des} = [{p}_\textnormal{y}^\textnormal{HLIP,des} \ {v}_\textnormal{y}^\textnormal{HLIP,des}]^\top$ are the desired x and y HLIP states.

Next, we convert the HLIP pre-impact references into ALIP pre-impact references the following way
\begin{equation}
\resizebox{\hsize}{!}{$
    \begin{bmatrix}
        {p}_\textnormal{x}^\textnormal{ALIP,des} \\
        {L}_\textnormal{y}^\textnormal{ALIP,des} \\
        {p}_\textnormal{y}^\textnormal{ALIP,des} \\
        {L}_\textnormal{x}^\textnormal{ALIP,des} 
    \end{bmatrix}
    =
    \begin{bmatrix}
        1 & 0 & 0 & 0 \\
        0 &m_\textnormal{COM} p_\textnormal{z}^\textnormal{des} & 0 & 0 \\
        0 & 0 & 1 & 0 \\
        0 & 0 & 0 & -m_\textnormal{COM} p_\textnormal{z}^\textnormal{des} 
    \end{bmatrix}
    \begin{bmatrix}
        {p}_\textnormal{x}^\textnormal{HLIP,des} \\
        {v}_\textnormal{x}^\textnormal{HLIP,des} \\
        {p}_\textnormal{y}^\textnormal{HLIP,des} \\
        {v}_\textnormal{y}^\textnormal{HLIP,des} 
    \end{bmatrix}.
$}
\end{equation}

\subsubsection{Obtaining the current state}
When solving the nonlinear MPC problem in \eqref{eq:nonlinear_mpc}, we set $\mathbf{x}_0^\textnormal{ALIP}$ to the current ALIP state of the robot. The current position state of the ALIP model can be directly calculated based on the current COM position relative to the stance foot. The current angular momentum in the sagittal and frontal plane we approximate based on the current state as
\begin{align}
    \label{eq:l_y_ref}
    L_y &= m_\textnormal{COM} p_\textnormal{z} v_\textnormal{x} - m_\textnormal{COM} p_\textnormal{x} v_\textnormal{z} + I_\textnormal{COM,y} \omega_\textnormal{y},  \\
    \label{eq:l_x_state}
    L_x &=-m_\textnormal{COM} p_\textnormal{z} v_\textnormal{y} + m_\textnormal{COM} p_\textnormal{y} v_\textnormal{z} + I_\textnormal{COM,x} \omega_\textnormal{x}.
\end{align}

Where we assume that the inertial matrix is diagonal, and that $I_\textnormal{COM,x}$ and $I_\textnormal{COM,y}$ correspond to the inertia about the robot's COM around the x and y-axis. 

\subsection{SRB Reference Generation}\label{sec:method:reference_generation}

\subsubsection{S2S References}

Solving the NMPC from \eqref{eq:nonlinear_mpc} yields the following S2S trajectories: $\hat{\bar{\mathbf{X}}}^\textnormal{ALIP} = \left\{ \bar{\mathbf{x}}_k^\textnormal{ALIP} \right \}_{k = 1}^{K}$, \newline$\hat{\bar{\mathbf{L}}} = \left \{\bar{\boldsymbol{\ell}}_k \right \}_{k=0}^{K-1}$, $\hat{\bar{\mathbf{T}}} = \left\{ \bar{{T}}_k \right \}_{k=0}^{K-1}$, and $\hat{\bar{\boldsymbol{\tau}}}= \left\{ \bar{\boldsymbol{\tau}}_k \right \}_{k=0}^{K-1}$. 

Next, by using the commanded turning rate, $\omega_\textnormal{z}^\textnormal{cmd}$, we can obtain the stance foot yaw orientation for each step relative to the current stance foot yaw orientation $\hat{\boldsymbol{\Psi}} = \left \{ \psi_k^\textnormal{STF} \right \}_{k=0}^{K-1}$.

Using the S2S solution trajectories, we can calculate the robot's stance foot position at each step $k$, relative to the current stance foot position, $\mathbf{p}^\textnormal{STF}_0 = \mathbf{0}$, i.e. the inertial frame:
\begin{equation}
    \label{eq:stf_position}
    \bar{\mathbf{p}}_{\textnormal{xy,}k}^\textnormal{STF} = \sum_{i=1}^{k} \mathbf{R}_\textnormal{z}(\psi_{i-1}^\textnormal{STF}) \boldsymbol{\ell}_i, \quad k \in \{1, \dots, K{-}1\} 
\end{equation}
In \eqref{eq:stf_position}, $\mathbf{R}_\textnormal{z}$ denotes a simple rotation around the z-axis.

\subsubsection{SRB References}
Let the time of node $n$ be given by $t_n$. Each node $n$ is associated with a step $k_n$, whose pre-impact time is given by $t_{k_n}^\textnormal{impact}$. We then have that the time-until-impact (T2I) for node $n$ is given by $t_{n}^\textnormal{T2I} = t_{k_n}^\textnormal{impact} - t_n$. To obtain the desired ALIP state at node $n$, we flow the continuous ALIP dynamics from \eqref{eq:nonlinear_forward_propagation}  backward in time from the pre-impact state as follows:
\begin{equation}
    \label{eq:nonlinear_backward_propagation}
    \bar{\mathbf{x}}^\textnormal{ALIP}_n = \boldsymbol{\Phi}(-t_n^\textnormal{T2I})  \bar{\mathbf{x}}^\textnormal{ALIP}_{k_n} + \boldsymbol{\Gamma}(-t_n^\textnormal{T2I}) \boldsymbol{\tau}_{k_n}.
\end{equation}
The COM xy position relative to the current stance foot frame can be obtained through the following transformation:
\begin{equation}
    \bar{\mathbf{p}}_{\textnormal{xy,}n}^\textnormal{COM} = \bar{\mathbf{p}}_{k_n}^\textnormal{STF} + \mathbf{R}_\textnormal{z}(\psi_{k_n})       \bar{\mathbf{p}}_{\textnormal{xy},n}^\textnormal{ALIP},
\end{equation}
and the reference velocity at node $n$ is given by 
\begin{equation}
    \bar{\mathbf{v}}^\textnormal{COM}_{\textnormal{xy},n} = \mathbf{R}_\textnormal{z}(\psi_{k_n}^\textnormal{STF}) \bar{\mathbf{v}}_{\textnormal{xy,} k_n}^\textnormal{HLIP}.
\end{equation}
We then compute the SRB reference at node $n$ as follows:
\begin{equation}
\label{eq:x_srb_ref}
\scalebox{0.95}{$
    \bar{\mathbf{x}}_n^\textnormal{SRB}
    = 
    \begin{bmatrix}
        (\bar{\mathbf{p}}_{\textnormal{xy,}n}^\textnormal{COM})^\top \
        p_\textnormal{z}^\textnormal{des} \
        \bar{\phi} \
        \bar{\theta} \
        \bar{\psi}_n^\textnormal{COM} \
        (\bar{\mathbf{v}}_{\textnormal{xy,}n}^\textnormal{COM})^\top \
        \mathbf{0}_{3 \times 1}\ \
        \bar{\omega}_{\textnormal{z,} n}^\textnormal{cmd} \
        g
    \end{bmatrix}^\top
$}
\end{equation}

\begin{figure*}[t]
\begin{equation}
\label{eq:dsrb_dynamics}
\resizebox{0.94\textwidth}{!}{$
\begin{aligned}
\begin{bmatrix}
\dot{\mathbf{x}}^\textnormal{SRB} \\
\dot{\psi}_\textnormal{TR} \\
\dot{p}_\textnormal{LA} \\
\dot{p}_\textnormal{RA} \\
\ddot{\psi}_\textnormal{TR} \\
\dot{v}_\textnormal{LA} \\
\dot{v}_\textnormal{RA}
\end{bmatrix}
&=
\underbrace{
\begin{bmatrix}
\begin{array}{c|ccc|c}
\multirow{4}{*}{$\mathbf{A}^\textnormal{SRB}$} 
& \mathbf{0}_{10 \times 1} & \mathbf{0}_{10 \times 1} & \mathbf{0}_{10 \times 1} & \mathbf{0}_{10 \times 3} \\
& 0 & m_\textnormal{LA} g & m_\textnormal{RA} g & \mathbf{0}_{1 \times 3} \\
& \mathbf{0}_{2 \times 1} & \mathbf{0}_{2 \times 1} & \mathbf{0}_{2 \times 1} & \mathbf{0}_{2 \times 3} \\
\hline
\mathbf{0}_{3 \times 13} & \mathbf{0}_{3 \times 1} & \mathbf{0}_{3 \times 1} & \mathbf{0}_{3 \times 1} & \boldsymbol{I}_{3 \times 3} \\
\mathbf{0}_{3 \times 13} & \mathbf{0}_{3 \times 1} & \mathbf{0}_{3 \times 1} & \mathbf{0}_{3 \times 1} & \mathbf{0}_{3 \times 3}
\end{array}
\end{bmatrix}
}_{\mathbf{A}^\textnormal{DSRB}}
\begin{bmatrix}
\mathbf{x}^\textnormal{SRB} \\
\psi_\textnormal{TR} \\
p_\textnormal{LA} \\
p_\textnormal{RA} \\
\dot{\psi}_\textnormal{TR} \\
v_\textnormal{LA} \\
v_\textnormal{RA}
\end{bmatrix}
+
\underbrace{
\begin{bmatrix}
\begin{array}{c|ccc}
\multirow{4}{*}{$\mathbf{B}^\textnormal{SRB*}$} 
& \mathbf{0}_{6 \times 1} & \mathbf{0}_{6 \times 1} & \mathbf{0}_{6 \times 1} \\
& \mathbf{0}_{3 \times 1} & \mathbf{e}_\textnormal{x} / m_\textnormal{COM} & \mathbf{e}_\textnormal{x} / m_\textnormal{COM} \\
& -\mathbf{e}_\textnormal{z}/I_\textnormal{LB,z} & \boldsymbol{\alpha}_\textnormal{LA} & \boldsymbol{\alpha}_\textnormal{RA} \\
& 0 & 0 & 0 \\
\hline
\mathbf{0}_{3 \times 12} & \mathbf{0}_{3 \times 1} & \mathbf{0}_{3 \times 1} & \mathbf{0}_{3 \times 1} \\
\mathbf{0}_{1 \times 12} & 1/I_\textnormal{TR,z} & -r_\textnormal{LA,y}/I_\textnormal{TR,z} & -r_\textnormal{RA,y}/I_\textnormal{TR,z} \\
\mathbf{0}_{1 \times 12} & 0 & -1/m_\textnormal{LA} & 0 \\
\mathbf{0}_{1 \times 12} & 0 & 0 & -1/m_\textnormal{RA}
\end{array}
\end{bmatrix}
}_{\mathbf{B}^\textnormal{DSRB}}
\begin{bmatrix}
\mathbf{u}^\textnormal{SRB} \\
\tau_\textnormal{TR} \\
F_\textnormal{LA,x} \\
F_\textnormal{RA,x}
\end{bmatrix}
\end{aligned}
$}
\end{equation}
\vspace{-10.0mm}
\end{figure*}

\subsection{Linear MPC Formulation}\label{sec:method:arm_mpc}
In the classical SRB-MPC formulation for legged systems, the input vector $\mathbf{u}^\textnormal{SRB}$ only contains forces and moments applied to the robot due to contact with external objects \cite{di2018dynamic, li2021force}. Here we expand upon the classical legged SRB-MPC framework, by also incorporating the arms and torso into the MPC formulation while still ensuring linear dynamics. 

\subsubsection{Decomposing the SRB Model}
\begin{figure}[t]
    \centering
    \includegraphics[width=1.0 \linewidth]{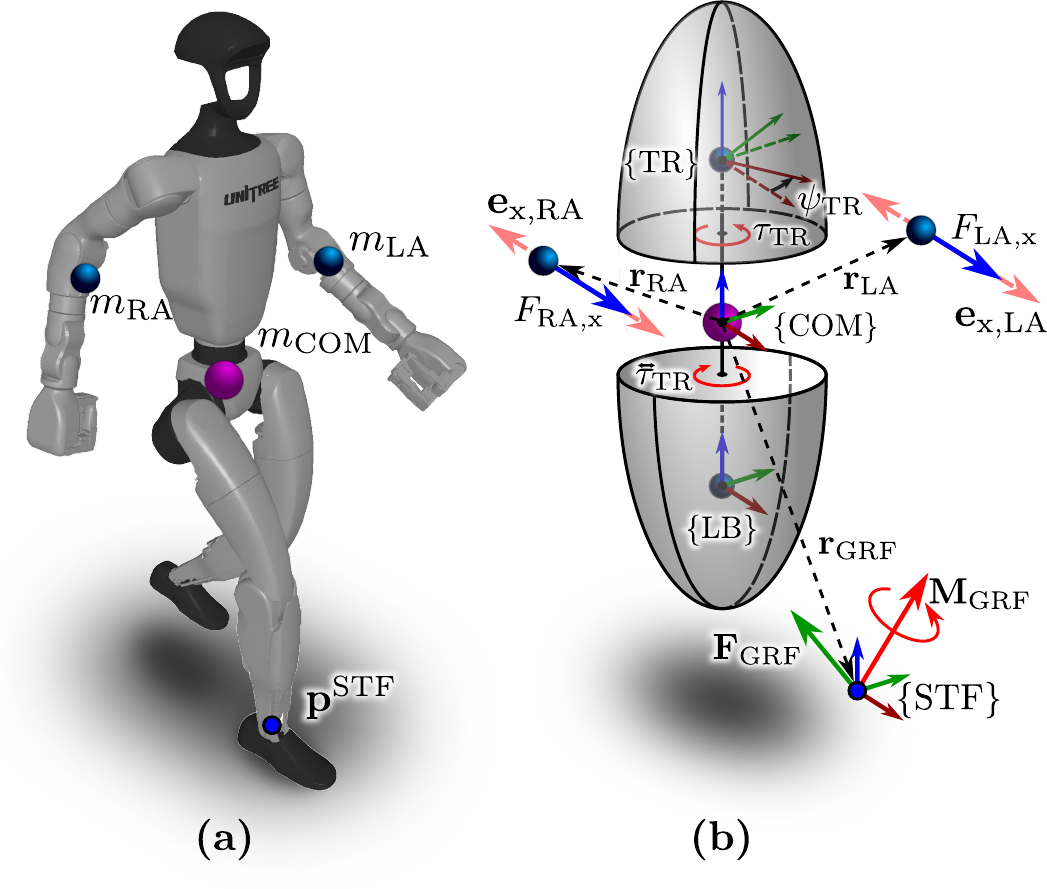}
    \vspace{-15pt}
    \caption{(a) The full humanoid robot model and (b) the proposed simplified model. We model the humanoid as four bodies: the lower body {(LB)}, the torso {(TR)}, and the left {(LA)} and right {(RA)} arm. The torso and lower body are connected through a torso z joint, and the arms are modeled as point mass that are connected to the torso. }
    \label{fig:whole_body_and_rom}
    \vspace{-6mm}
\end{figure}

We model the humanoid as four parts; the lower body {(LB)}, the torso {(TR)}, and the left {(LA)} and right {(RA)} arms. We model the left and right arm as point masses, $m_\textnormal{LA}$ and $m_\textnormal{RA}$, which are located at $\mathbf{r}_\textnormal{LA}$ and $\mathbf{r}_\textnormal{RA}$. We assume that the torso can rotate relative to the lower body via a z rotation, and that the arms are connected to the torso. In addition to the ground reaction forces and moments available in the standard SRB model, our proposed model also allows us to generate forces acting on the robot's torso and lower body by accelerating the arms, and a z moment pair can be generated between the lower body and the torso by applying a moment in the torso z actuator. Figure~\ref{fig:whole_body_and_rom} illustrates the proposed model. 

\begin{remark} While we refer to this as a decomposition of the SRB model, it can also be viewed as a simplified or coarse centroidal model \cite{dai2014whole}. Unlike the standard SRB \cite{di2018dynamic}, our model separates the robot into multiple sub-bodies, providing a richer representation of internal force generation while maintaining computational simplicity compared to full centroidal dynamics. 
\end{remark}

\subsubsection{Dynamics}


If we accelerate one of the arm masses, e.g. the left arm, relative to the COM by $\mathbf{a}_\textnormal{LA} \in \mathbb{R}^3$, we effectively apply a force $\dvec{\mathbf{F}}_\textnormal{LA} = m_\textnormal{LA} \mathbf{a}_\textnormal{LA}$ to the arm and in return an equal and opposite force $\mathbf{F}_\textnormal{LA} = - \dvec{\mathbf{F}}_\textnormal{LA}$ is applied to the robot at $\mathbf{r}_\textnormal{LA}$. When applying ${\mathbf{F}}_\textnormal{A}$ at $\mathbf{r}_\textnormal{LA}$, the following forces and moments are applied to the rest of the robot
\begin{equation}
    \label{eq:arm_dynamics_on_robot}
    \begin{bmatrix}
        \mathbf{F}_\textnormal{LA} \\
        \mathbf{M}_\textnormal{LA}
    \end{bmatrix}
    = 
    \begin{bmatrix}
        \mathbf{F}_\textnormal{LA} \\
        (\mathbf{r}_\textnormal{LA})^{\times} \mathbf{F}_\textnormal{LA}
    \end{bmatrix}.
\end{equation}
From \eqref{eq:arm_dynamics_on_robot} we have that the moments generated by the arms on the robot are nonlinear due to the coupling between the $\mathbf{r}_\textnormal{LA}$ vector and the applied force $\mathbf{F}_\textnormal{LA}$, which is unsuitable for linear MPC. However, \eqref{eq:arm_dynamics_on_robot} can be made linear with respect to the forces and arm positions if we constrain the motion of the arm mass to move along a single axis. 

In this paper we are mainly interested in using the arms to stabilize the yaw of the robot's base, i.e. the pelvis, hence we choose to only allow movement of the arm mass along the x-axis, $\mathbf{e}_\textnormal{x,LA}$. We therefore allow $r_\textnormal{LA,x}$ to move, while we keep $r_\textnormal{LA,y}$ and $r_\textnormal{LA,z}$ fixed. Furthermore, to keep the arm movement fixed in the y and z-direction, we also have that $F_\textnormal{LA,y} = 0$ and $F_\textnormal{LA,z} = 0$. With these constraints enforced we have that the arm forces simplify \eqref{eq:arm_dynamics_on_robot} into the following forces and moments acting on the robot

\begin{equation}
    \label{eq:arm_dynamics_on_robot_simple}
    \mathbf{u}_\textnormal{LA}
    =
    \begin{bmatrix}
        F_\textnormal{LA,x} \\
        M_\textnormal{LA,y} \\
        M_\textnormal{LA,z}
    \end{bmatrix}
    = 
    \begin{bmatrix}
        F_\textnormal{LA,x} \\
        r_\textnormal{LA,z} F_\textnormal{LA,x} \\
        - r_\textnormal{LA,y} F_\textnormal{LA,x}\\
    \end{bmatrix}.
\end{equation}

In addition to the forces and moments generated from the arms, our proposed decomposition also allows us to rotate the torso relative to the lower body about the z-axis. We denote the torso yaw relative to the lower body as $\psi_\textnormal{TR}$. Furthermore, since the lower and upper body are connected via an actuator located in the lower body, we can apply a torque $\tau_\textnormal{TR}$ to the torso, which equivalently leads to an equal and opposite torque $\dvec{\tau}_\textnormal{TR} = -\tau_\textnormal{TR}$ being applied to the lower body. By limiting the arm motion to one direction and including the torso z actuation, we define the Decomposed SRB (DSRB) state and input as
\begin{align}
    \label{eq:DSRB_state}
    \mathbf{x}^\textnormal{DSRB} &= \left[ (\mathbf{x}^\textnormal{SRB})^\top \ 
    \psi_\textnormal{TR} \ 
    p_\textnormal{LA} \ 
    p_\textnormal{RA} \
    \dot{\psi}_\textnormal{TR} \
    v_\textnormal{LA} \
    v_\textnormal{RA}
    \right]^\top,\\
    \label{eq:DSRB_input}
    \mathbf{u}^\textnormal{DSRB} &= \left[ (\mathbf{u}^\textnormal{SRB})^\top \ 
    \tau_\textnormal{TR} \ 
    F_\textnormal{LA,x} \ 
    F_\textnormal{RA,x} 
    \right]^\top,
\end{align}
and the corresponding linear dynamics as shown in \eqref{eq:dsrb_dynamics}. In \eqref{eq:dsrb_dynamics}, $\boldsymbol{\alpha}_\textnormal{LA} = \left[ 0 \ \frac{r_\textnormal{LA,z}}{I_\textnormal{COM,y}} \ 0 \right]^\top$, $\boldsymbol{\alpha}_\textnormal{RA} = \left[ 0 \ \frac{r_\textnormal{RA,z}}{I_\textnormal{COM,y}} \ 0 \right]^\top$, $\mathbf{e}_\textnormal{x} = [1 \ 0 \ 0]^\top$, $\mathbf{e}_\textnormal{z} = [0 \ 0 \ 1]^\top$, and $I_\textnormal{LB,z}$ and $I_\textnormal{TR,z}$ is the inertia around the z-axis for the lower body and torso, respectively. Since the robot is divided into the lower body and torso, $I_\textnormal{COM,z}$ in $\mathbf{B}^\textnormal{SRB}$ is replaced by $I_\textnormal{LB,z}$ in $\mathbf{B}^\textnormal{SRB*}$.

Similarly as for the standard SRB dynamics from \eqref{eq:continous_time_dynamics}, we can discretize the DSRB dynamics and solve a linear optimization problem at the same form as in \eqref{eq:srb_mpc}.

\subsection{Low Level Control}\label{sec:method:low_level_control}

The joint position commands, $\mathbf{q}_\textnormal{p}^\textnormal{cmd}$, are calculated using inverse kinematics based on the current state and the DSRB trajectory. For the swing foot trajectory we use the method proposed in \cite{ghansah2024dynamic} to ensure smooth blending between the current state and the target state.

We calculate the feed-forward joint torques by using the Jacobian from the COM to the components of $\mathbf{u}^\textnormal{DSRB}$, together with the first control input from the MPC controller
\begin{equation}
    \label{eq:jacobian}
    \mathbf{q}_\tau^\textnormal{cmd} = \mathbf{J}^\top \mathbf{u}_0^\textnormal{DSRB}
\end{equation}
In \eqref{eq:jacobian} we expand $\mathbf{u}_0^\textnormal{DSRB}$ to also enforce that the arm forces in the y and z-direction equal zero, i.e. \newline $F_\textnormal{LA,y} = F_\textnormal{LA,z} = F_\textnormal{RA,y} = F_\textnormal{RA,z} \equiv 0$. Even though we optimize over ankle torques in the S2S ALIP dynamics, we do not pass those references directly to the robot, but instead use the values obtained from \eqref{eq:jacobian}.

\section{Experimental Results}\label{sec:experiments}

We validate our control framework through simulation and hardware experiments on the 29-DOF, 35 kg Unitree G1 humanoid robot. We first demonstrate through simulation experiments how incorporating the upper body in the mid-level MPC enhances the robot's ability to reject external yaw-moment disturbances. Next, in Section \ref{sec:experiments:step_frequency} we show how the adaptive step frequency improves the stability of the robot when being exposed to various external disturbances. Finally, in Section \ref{sec:experiments:hardware} we demonstrate the robustness of the robot through hardware experiments, where we show how the nonlinear step-planner modifies the step periods and step lengths when the robot is pushed.

\subsection{Implementation Details}\label{sec:experiments:setup}
The high level NMPC planner was implemented using Casadi's \cite{Andersson2019} C++ interface and the IPOPT solver \cite{wachter2002interior}. We solve the NMPC at a fixed frequency of 40 Hz. The rest of the controller, i.e. the trajectory generator, the mid-level MPC, and the low-level command mapper all run at 500 Hz. We utilize the Eigen library \cite{eigenweb} to perform linear algebra operations and Pinocchio \cite{carpentier2019pinocchio} to compute kinematics and dynamics. The simulation experiments were conducted using the MuJoCo physics engine \cite{todorov2012mujoco}. Both in simulation and on hardware we run a strap down inertial navigation system \cite{fossen1999guidance} at 500 Hz to estimate the base pose and twist state of the robot, where we receive IMU and joint measurements at 500 Hz, and pose and twist measurements at 200 Hz.

\begin{figure}[t]
    \centering
    \includegraphics[width=1.0\linewidth]{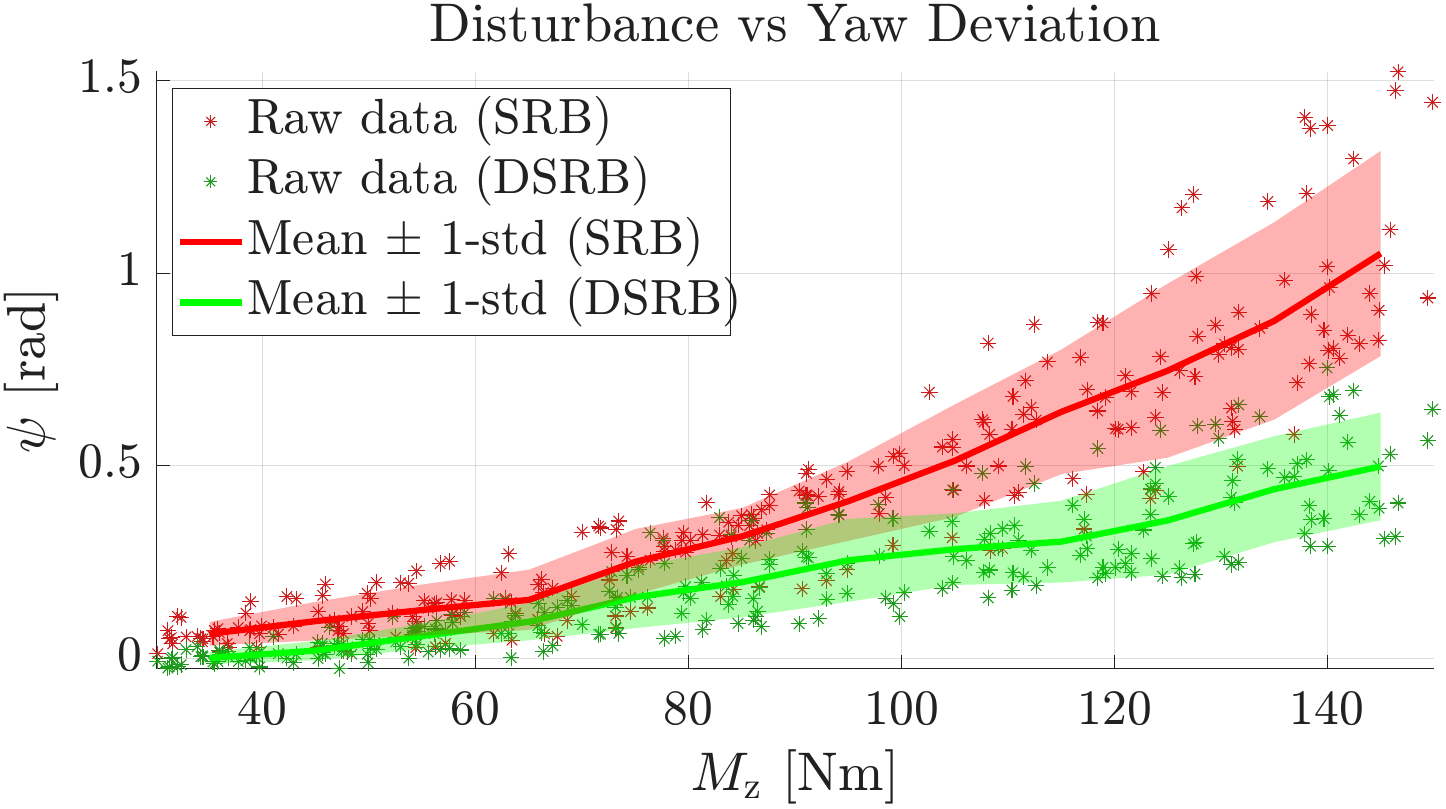}
    \caption{Pelvis yaw after recovery from torso yaw disturbances (30 -- 150 Nm for 0.05 s) when walking at 0.3 m/s. Red: SRB-MPC. Green: DSRB-MPC. Means and standard deviations are shown in 10 Nm bins}  
    \label{fig:yaw_disturbance_simulation}
    \vspace{-2mm}
\end{figure}

\begin{figure}[h]
    \centering
    \includegraphics[width=1.0\linewidth]{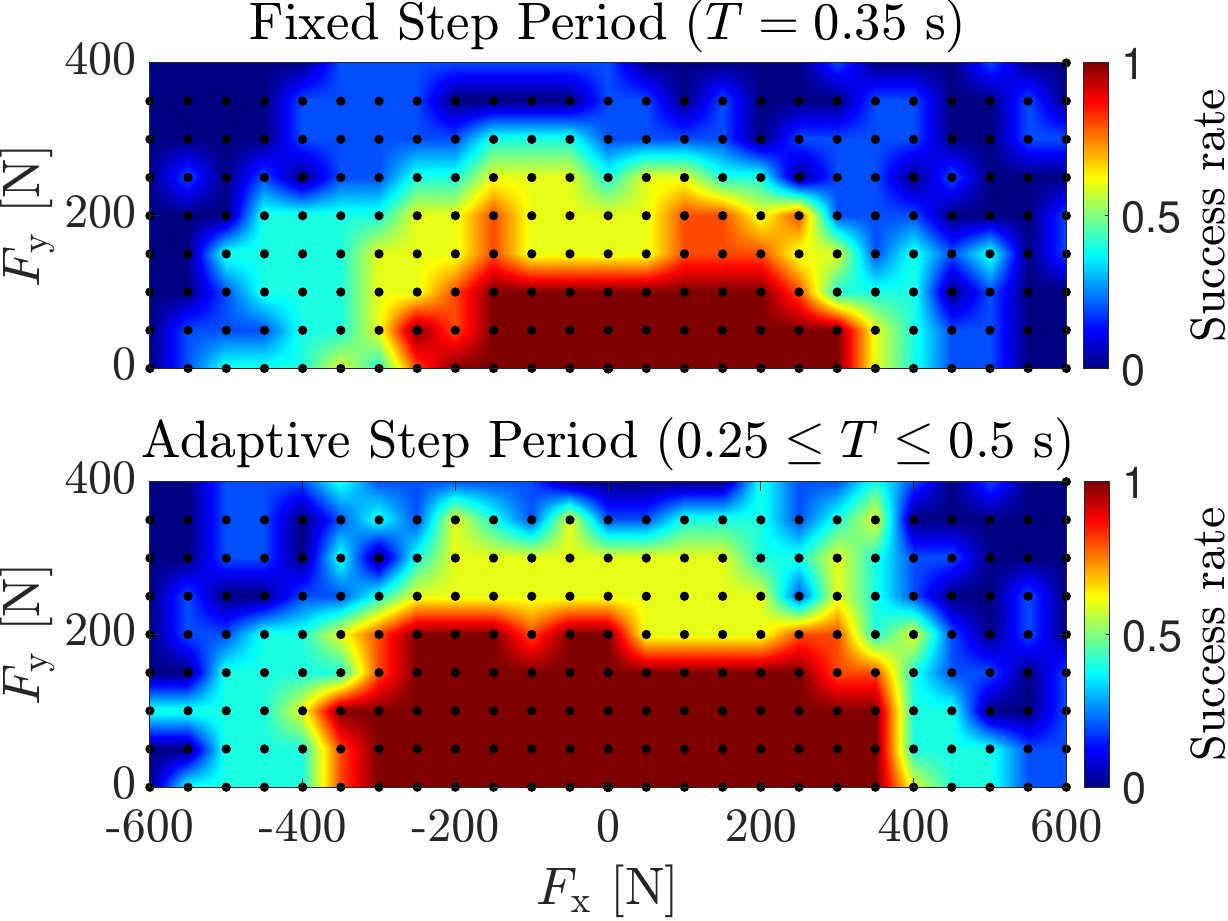}
    \caption{Success rates for push recovery under external disturbances in the x-direction (–600 -- 600 N) and y-direction (0 -- 400 N), each lasting 0.1 s. Top: fixed step period of 0.35 s. Bottom: adaptive step period in the range 0.25 -- 0.5 s. Color bar: success rate (1 = 100 \%).}
    \label{fig:step_frequency_simulation}
    \vspace{-6mm}
\end{figure}

\subsection{Upper Body-Assisted Disturbance Rejection}\label{sec:experiments:upper_body}

To demonstrate the effectiveness of our proposed mid-level DSRB controller over the nominal SRB controller, we first make the robot walk forward at a fixed speed of 0.3 m/s. In each test we apply a $M_\textnormal{z}$ moment in the range 30 -- 150 Nm for 0.05 s to the robot's torso, and then after the robot is stabilized, we record the yaw of the base, i.e., the pelvis. As shown in Fig. \ref{fig:yaw_disturbance_simulation}, the DSRB-MPC consistently reduces yaw deviations over the SRB-MPC, where the difference in error on average grows with the magnitude of the yaw disturbances. When incorporating upper body control the robot can more effectively reject external yaw moments, as torso moment and arm forces can be used to mitigate pelvis rotation. When disabling the upper body control, the robot relies more on the friction from the stance foot.

\begin{figure*}[t]
    \centering
    \includegraphics[width=0.95\linewidth]{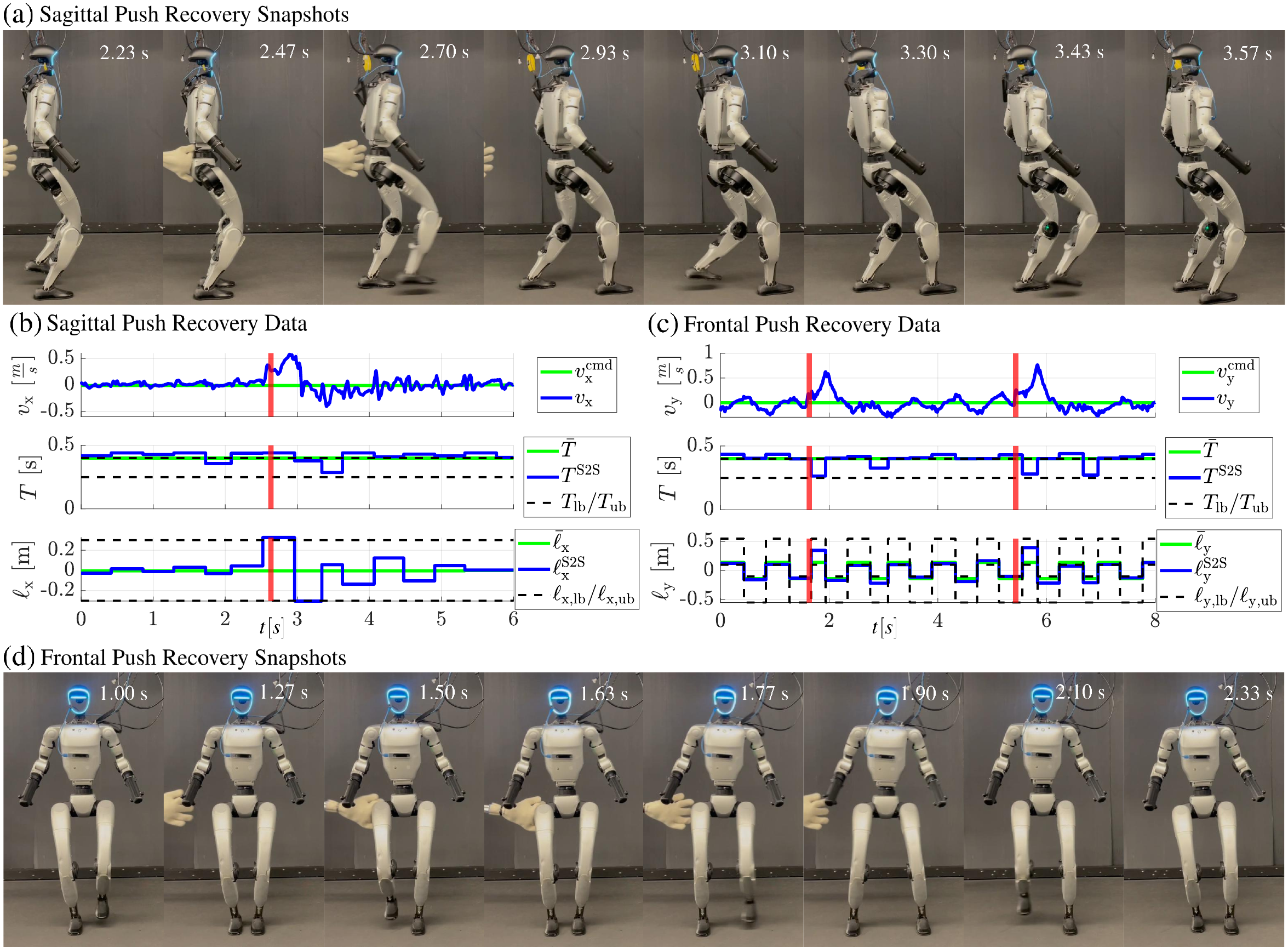}
    \caption{The robot is exposed to external disturbances while walking in place. The plots shows how the stepping periods and step lengths change to stabilize the robot. In (a) and (b) the robot is pushed in the x-direction, and (c) and (d) the robot is pushed in the y-direction. Here the S2S superscripts refer to the final step lengths and step period of the robot at each step.}
    \label{fig:step_frequency}
    \vspace{-6mm}
\end{figure*}

\subsection{Adaptive Step-Frequency Stabilization}\label{sec:experiments:step_frequency}

We evaluated the benefit of optimizing over step periods \textit{and} step lengths, as opposed to only step lengths, by performing push recovery tests. In each test, the robot was walking in place and then an external force ($F_\textnormal{x}$: –600 -- 600~N, $F_\textnormal{y}$: 0 -- 400~N) was applied for 0.1 s. If the robot was still walking after being pushed the test was deemed a success. For each force combination we performed the test 5 times, where the force was applied at different times to account for varying step phases. Figure \ref{fig:step_frequency_simulation} (top) visualizes the results when using a fixed stepping period of 0.35 s, and Fig. \ref{fig:step_frequency_simulation} (bottom) when using an adaptive step period in the range 0.25 -- 0.5 s, with a desired step period of 0.4 s. Step period adaptation increased the number of successful recoveries by 36~\%, despite having a lower average step frequency, and the controller withstood larger disturbances in the x and y-direction. The step adaptation allows the NMPC to search over a larger set of potential solutions to stabilize the robot. In particular, when the robot is pushed, it is often advantageous to step quickly to recover faster.

\subsection{Hardware Validation}\label{sec:experiments:hardware}

\subsubsection{Hardware Setup}

The hardware experiments were run using a Minisforum EM780 computer, a low-power small-scale computer equipped with an AMD Ryzen 7 7840U CPU, and 32 GB of RAM. The onboard computer was powered by a power bank strapped to the robot, and communicated with the robot's internal computer over Ethernet. The onboard computer received joint state and IMU measurements from the robot, and pose and twist measurements from an onboard Intel RealSense T265 camera mounted to the torso.

\subsubsection{Push Recovery Experiments}

To investigate the robustness of our proposed walking controller we conducted multiple push recovery experiments. Two of these experiments are presented in Fig. \ref{fig:step_frequency}. In these experiments the robot was walking in place and then suddenly subjected to an external push force. In Fig. \ref{fig:step_frequency}a) the disturbance is acting in the x-direction, and in \ref{fig:step_frequency}d) the disturbance is acting in the y-direction. The corresponding plots, Fig. \ref{fig:step_frequency}b) and Fig. \ref{fig:step_frequency}c), show how the base velocity, step periods, and step lengths adaptively change in the two experiments. 

In the sagittal push recovery experiment we see that after the push occurs at $t = 2.63$~s, the velocity in the x-direction increases quickly. We can see that the controller responds by taking a couple of long steps to stabilize the robot rapidly. This can also be seen from snapshot $t=$2.47 -- 3.30 s in Fig. \ref{fig:step_frequency}a) where the robot takes two long steps. 

In the frontal push recovery experiments the robot is exposed to two forces in the y-direction, first at $t = 1.63$~s and then at $t = 5.43$~s. In both cases the y-velocity quickly increases after the pushes and the robot immediately takes longer and faster steps to stabilize itself, before going back to its nominal stepping pattern. 

\subsubsection{Locomotion Across Diverse Terrains}

In addition to the push recovery experiments we also conducted various other walking experiments. These experiments included walking over rough terrain indoors (Fig. \ref{fig:hero}c), blindly stepping down from large heights (Fig. \ref{fig:hero}d), outdoor walking on stone pavement (Fig. \ref{fig:hero}b), and walking on grass (Fig. \ref{fig:hero}a). In these hardware experiments we disabled the upper body control.

\begin{remark}
Despite relying on simplified ROMs that assume a point-mass or fixed inertia, our hierarchical control framework robustly stabilizes the whole-body dynamics of the humanoid, including non-negligible limb inertias. This demonstrates the effectiveness and generalizability of the approach, even when key modeling assumptions are violated.
\end{remark}

\section{Conclusion}\label{sec:conclusion}

In this paper, we presented a hierarchical control framework for robust humanoid locomotion based on reduced-order models. The high-level planner utilizes NMPC on the ALIP model to jointly optimize over step periods and step lengths, and its S2S plan is converted into an SRB trajectory tracked by a linear MPC. We proposed a decomposed SRB model that incorporates arm and torso dynamics while maintaining linear constraints in the MPC. The method was demonstrated through simulation and hardware experiments, showing that upper-body control improves yaw disturbance rejection and that step adaptation increases robustness.

A limitation of the current framework is that the high-level ALIP model does not capture the richer torso and arm dynamics represented in the DSRB model. Future work will explore closer integration of the planners and extensions to more dynamic gaits such as running and hopping.


\balance
\bibliographystyle{IEEEtran}

\bibliography{99_references}


\end{document}